\documentclass[a4paper]{article}

\usepackage{INTERSPEECH2019}
\usepackage{booktabs}
\usepackage{multirow}
\usepackage{siunitx}

\title{Visualization and Interpretation of Latent Spaces for Controlling Expressive Speech Synthesis through Audio Analysis}
\name{No\'e Tits$^1$, Fengna Wang$^2$, Kevin El Haddad$^1$, Vincent Pagel$^2$, Thierry Dutoit$^1$}

\address{
  $^1$Numediart Institute, University of Mons\\
  $^2$Acapela Group}

\email{\{noe.tits, kevin.elhaddad, thierry.dutoit\}@umons.ac.be,\\
\{fengna.wang, vincent.pagel\}@acapela-group.com }

\begin{document}

\maketitle
\begin{abstract}

The field of Text-to-Speech has experienced huge improvements last years benefiting from deep learning techniques. Producing realistic speech becomes possible now. As a consequence, the research on the control of the expressiveness, allowing to generate speech in different styles or manners, has attracted increasing attention lately. Systems able to control style have been developed and show impressive results. However the control parameters often consist of latent variables and remain complex to interpret.

In this paper, we analyze and compare different latent spaces and obtain an interpretation of their influence on expressive speech. This will enable the possibility to build controllable speech synthesis systems with an understandable behaviour.

\end{abstract}

\section{Introduction}

During the last few years, many Text-to-Speech systems based on Deep Learning were developed and showed remarkable performance in terms of reliability and quality of speech. Lately, researchers in this area have been focusing on controlling the speech variability of this kind of systems~\cite{generative_controllable_speech-18-hsu,unsupervised_controllable_speech-18-henter,style_tokens-18-wang}.

An issue for such a control is the lack of data labeled with information such as emotion or style. Emotion modeling is thus one of the main challenges in developing more natural human-machine interfaces. Two main approaches exist to modeling emotions. 

A first representation is the categorical representation, such as Ekman's six basic emotion model~\cite{basic_emotions-92-Ekman} which identify anger, disgust, fear, happiness, sadness and surprise as six basic emotions from which the other emotions may be derived.
Some speech datasets~\cite{ravdess18emoDB,cremad14emoDB,emov-db-18-tits, berlinEmo05emoDB} are annotated in emotional categories. This kind of datasets allow the development of category based emotional TTS~\cite{exploring_transfer_learning-19-tits, emo_tacotron-17-lee}.
The disadvantage of such simple annotation is that they do not offer a continuous representation of emotion.

Emotions can also be represented in a multidimensional continuous space like in the Russels circumplex model~\cite{circumplex-80-russel}. This modeling allows to better reflect the complexity and the variations in the expressions, unlike the category system. The two most commonly used dimensions in the literature are the arousal corresponding to the level of excitation and the valence corresponding to the pleasure level or positiveness of the emotion.
In datasets~\cite{improv-17-busso,iemocap-08-busso} annotated in emotional dimensions, for each utterance, the final emotion value were obtained by averaging over all annotated results from raters. However they are not suitable for synthesis purpose because they contain overlapping speech due to the data recording setup (dyadic conversation) and some external noise. Moreover, humans are not reliable for giving absolute values to estimate subjective emotional variables~\cite{miller1956magical}.

Some researchers tackled the problem of how to capture emotional representation by training systems on other tasks, leading to different approaches employing transfer learning techniques~\cite{asr-based-features-18-tits}.

Recent researches have proposed unsupervised techniques to achieve controllable speech synthesis, avoiding the problem of labels.

In~\cite{tacotron_prosody-18-skerry}, the authors present an extension to the Tacotron speech synthesis architecture that learns a latent embedding space by encoding audio into a vector that conditions Tacotron along with the text representation. These latent embeddings model the remaining variation in speech signals after accounting for variation due to phonetics, speaker identity, and channel effects.

In~\cite{expressive_speech_vae-18-akuzawa}, the Variational Auto-encoder (VAE) is used in a speech synthesis system, in combination with VoiceLoop.

Some other researches have used the concept of VAE~\cite{generative_controllable_speech-18-hsu,unsupervised_controllable_speech-18-henter,style_tokens-18-wang}. In~\cite{generative_controllable_speech-18-hsu}, they % who is they? (voir partout) -> alors surtout google, parfois des "IEEE members"
combine VAE with Gaussion Mixture Model (GMM) and call it GMVAE. In~\cite{unsupervised_controllable_speech-18-henter}, they use Vector Quantization with a VAE (VQ-VAE).

These works show that is is possible to build a latent space leading to variables that can be used to control style in speech synthesis. In~\cite{generative_controllable_speech-18-hsu}, they show that their system can generate spectrograms with different rythm, speaking rate and $F_0$ from a single text.
% je ne vois pas le lien entre cette dernière phrase et ce qui précède.

However these works do not provide insights about the relationships between the resulting latent space and the audio characteristics that are possible to control.

In this paper, we aim to build latent spaces from an audio dataset containing different speech styles. We want the latent spaces to be useful to control a speech synthesis system. We use classical feature selection techniques as a way to compare various embedding types to evaluate their ability to discriminate between styles.  We then study relationships between each latent space and audio features to inspect the remaining variability that could be used to control in speech generation. 

For this purpose, we compare three latent spaces computed by training deep learning-based systems on three different tasks:

\begin{itemize}
    \item Style classification
    \item Speaker classification
    \item Text-to-Speech with a VAE
\end{itemize}

\section{Dataset Used}
\label{dataset}
The dataset used in this work is a proprietary dataset of Acapela Group SA. 
The goal of these recordings was to build a storytelling system to build audiobooks from transcripts.

It contains phonetically rich sentences uttered by a male actor in English. The actor was asked to utter a set of the sentences in 8 style classes. For the sake of clarity, we will refer to Will in the following sections to designate the actor.

The instructions/examples given to Will to speak with different styles are the following:
\begin{itemize}
    \setlength\itemsep{0.1em}
    \item neutral: classical narration
    \item happy: smile and positive
    \item sad: depressed 
    \item bad guy: mean
    \item from afar: "open the gate!" said the knight
    \item proxy: "don't make too much noise or the monster will hear you", whispering
    \item old man: mimic an old man's voice
    \item little creature: little monster
    
    % \item bad guy: mean
    % \item from afar: "open the gate!" said the knight
    % \item happy: smile and positive
    % \item little creature: little monster
    % \item neutral: classical narration
    % \item old man: mimic an old man's voice
    % \item sad: depressed 
    % \item proxy: "don't make too much noise or the monster will hear you", whispering
\end{itemize}

For each sentence, there is a wave file sampled at 22.05 kHz and coded in 16 bit linear and a corresponding transcription. The duration of audio files are given in Table~\ref{db_durations} in minutes. The durations after trimming silences are also indicated.

\begin{table}[ht]
\caption{Durations (min), duration after trimming silences (min) an number of utterances for each style}
\label{db_durations}
\vspace{-5mm}
\begin{center}
\begin{tabular}{|c|c|c|c|}
\hline
{} &  Duration &  Trimmed duration &  n utts \\
\hline

NEUTRAL        &    240.68 &            150.50 &    3299 \\
HAPPY          &    152.00 &             97.08 &    2130 \\
SAD            &    199.02 &            142.20 &    2130 \\
BADGUY         &    179.74 &            113.57 &    1867 \\
FROMAFAR       &    190.02 &            119.14 &    2130 \\
PROXY        &    179.37 &            123.86 &    2232 \\
OLDMAN         &    239.51 &            134.38 &    2130 \\
LITTLECREATURE &    214.41 &            124.14 &    2156 \\

% BADGUY         &    179.74 &            113.57 &    1867 \\
% FROMAFAR       &    190.02 &            119.14 &    2130 \\
% HAPPY          &    152.00 &             97.08 &    2130 \\
% LITTLECREATURE &    214.41 &            124.14 &    2156 \\
% NEUTRAL        &    240.68 &            150.50 &    3299 \\
% OLDMAN         &    239.51 &            134.38 &    2130 \\
% SAD            &    199.02 &            142.20 &    2130 \\
% PROXY        &    179.37 &            123.86 &    2232 \\
\hline
\end{tabular}
\end{center}
\vspace{-5mm}
\end{table}

%(\textbf{Fengna: personally I think it's better that the order of styles in the table or in the text should be the same.})

\section{Embedding Computation Systems}
\label{embedding_computation}
\begin{figure}[ht]
  \centering
  \includegraphics[width=\linewidth]{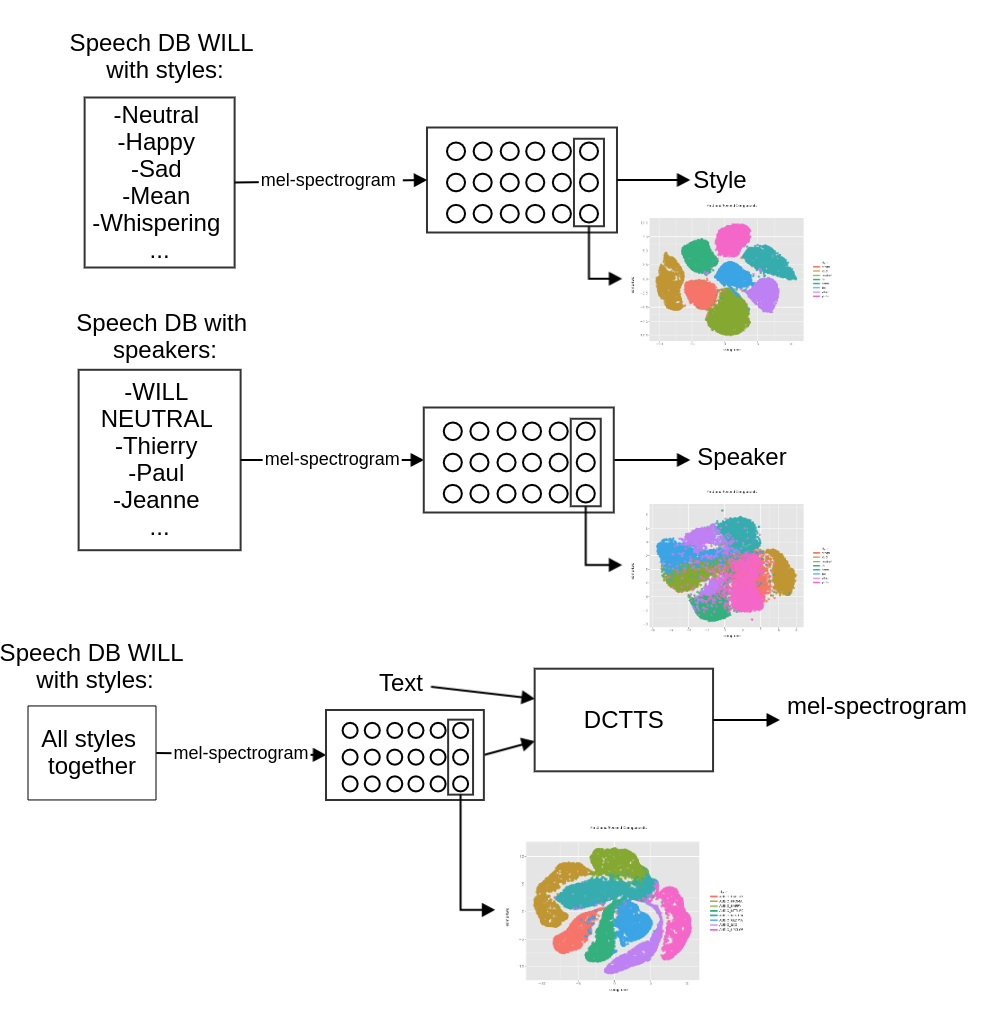}
  \caption{Illustration of data flow for the three embeddings computations systems. From top to bottom: Style classification system, speaker classification system and VAE-TTS.}
  \label{fig:embeddings_comparison}
\end{figure}

In this section, we describe the workflow of embedding computation. The three tasks used to generate embeddings are: Style Classification, Speaker Classification and VAE-TTS. Figure~\ref{fig:embeddings_comparison} illustrates the use of data to train the different systems. The colored point clouds illustrate the result of a dimension reduction performed on embeddings of Will dataset utterances. The colors correspond to styles of speech. Only the first system uses style labels during training. One can observed that points corresponding to styles are grouped together even for the two other systems in which style labels are not used during training.

To make a fair comparison among the three tasks, we restrict for each task the resulting embedding consisting of a 8-dimensional vector.  Better performance could be obtained with higher dimensional embeddings, like 128 dimensions in~\cite{tacotron_prosody-18-skerry} and 256 dimensions in~\cite{tacotron3-18-ye}, but it will add difficulties to analyze the relationship between embeddings and the audio characteristics that are possible to control in a controllable expressive speech synthesis system.

\subsection{Style Classification System}
\label{style_classif}
The classification system is a LSTM based DNN trained to predict the style category from audio features.
The DNN consists of three LSTM layers (512/128/64 cells in each layer) and a fully connected 8-dim embedding layer. The input is 80-bin mel-spectrogram of utterances.

The generated embeddings capture discriminative information specific to the class it belongs to, in our case, the classes refer to the 8 styles of Will voice.

Similar to~\cite{tacotron3-18-ye}, training utterances are firstly segmented into fixed % parle de chunks, pas de lenght chunks , partout -> 
length chunks without silence and the embedding from the last frame is taken as the style embedding of the considered chunk. 

In experiments, we tried different length chunks and found out that 800 ms gave the best classification performance. In the evaluation stage, we feed the whole utterance to the DNN to get the corresponding style embedding. %\textbf{Fengna: As we found out that in practice, the embedding from an utterance gave a slightly better classification accuracy than the averaged embedding from multiple fixed 800 ms chunks of the utterance.} 
Different from the speaker verification model in~\cite{generalized_loss_speaker-18-wan}, our style classifier uses style labels directly to compute loss without calculating the cosine similarity matrix. As our interest is to investigate latent embeddings that help to design an expressive speech synthesis system, the top priority is to having embeddings that contain useful emotional or style information. %Accordingly, style classification accuracy becomes the second target.

We obtained an accuracy of 94.38\% on style classification in our evaluation stage on the 800 unseen utterances (100 utterance per style). 
%Fengna: While 97.25\% classification accuracy with 128-dim embeddings.

\subsection{Speaker Classification System}
\label{speaker_classif}
As in Section~\ref{style_classif}, the speaker classification system is also a LSTM based DNN, but with four stacking LSTM with 512/512/512/64 cells in each layer, the 8-dim embedding layer is unchanged. 

In this case, the system is trained to predict the speaker identity from audio features. The speaker classifier is trained with 276 speakers' voices including the neutral subset of Will voice, the other styles of Will are not used during training. 

The 276 speakers are composed of speakers of varying ages (from child to the elder), female/male, and varying personalities. Most speakers speak in a neutral way, but some of them speak in a quite unique manner. Moreover, the 276 speakers cover 35 languages, with half of them speaking English with various accents, like American English, UK English, Austrian English, etc. The second most speaking language is French.
We would expect the embeddings generated from such a speaker classifier to reflect information not only about language, age, or gender , but also on prosody. As for the speaker classification system, utterances are firstly segmented into 800 ms chunks without silences in the training stage, and the entire utterance is fed to the classifier to get its embedding vector from the last non-silent frame. 
To train the classifier, we used as the training set 300 utterances per speaker and 100 utterances per speaker as the test set. Eventually we obtained 91.19\% classification accuracy on the unseen test utterances.

\subsection{VAE-TTS System}
\label{tts}
% Définir VAE, et expliquer quelles parties modélisent quelles distributions de probabilités
% Dire qu'il y a un average pooling après 
% Details of architecture

The TTS system is a deep learning algorithm trained to predict a spectrogram from the associated text input.

The TTS system used in this work is DCTTS~\cite{dctts-17-tachibana}. DCTTS models a sequence-to-sequence task with a encoder-decoder structure coupled with an Attention Mechanism like Tacotron~\cite{tacotron-17-wang}. Contrary to Tacotron, the modules of the architecture do not contain any recurrent unit. It is only based on convolutional modules. This particularity makes it easier to train. In ~\cite{dctts-17-tachibana}, they compared an open source implementation of Tacotron~\footnote{https://github.com/keithito/tacotron} to DCTTS and report higher Mean Opinion Score~(MOS).

In this work, we use the Tensorflow implementation available online~\footnote{https://github.com/Kyubyong/dc\_tts}.

There are two modules trained separately: Text2Mel and SSRN (for Spectrogram Super-resolution Network).
Text2Mel does the mapping between character embeddings and the output of Mel Filter Banks (MFBs) applied on the audio signal, that is, a mel-spectrogram. Then the second module SSRN does the mapping between the mel-spectrogram and full resolution spectrogram. Finally, Griffin-Lim~\cite{griffin_lim-84-griffin} is used as a vocoder.

Text2Mel module models the sequence-to-sequence task. It is composed of a Text  Encoder,  an Audio  Encoder,  an Attention Mechanism,  and  an Audio  Decoder.

In this paper we built a extension of DCTTS similar to the extension of Tacotron described in~\cite{tacotron_prosody-18-skerry}.

We encode the mel-spectrogram into a vector and concatenate this vector to each character of the transcript embedding and train the system for TTS. We chose a size of 8 to be relatively small compared to mel-spectrogram size (80 bins). The goal is to have a bottleneck sufficiently narrow to avoid the network to learn to copy the input at output.

The embeddings are therefore trained to represent the remaining variance in audio that does not depend on text. There is no need for labels. This enables the use of audiobooks not annotated in "style" or "emotion" for training expressive speech synthesis systems.

\section{Audio Analysis and Interpretation of Latent Spaces}

Latent spaces computed in Section~\ref{embedding_computation} should be useful to control a speech synthesis system. The goal is thus to have a latent space that has interpretable relationships with audio features that we could use to control in speech generation.
% tu ne parles pas de 4.1...

To that end, in Section~\ref{score} we evaluated the suitability of three embeddings for style classification. A good embedding must perform well in the classification task. Then we looked into the latent spaces in a closer detail.

This is investigated in Section~\ref{correlations} through an analysis of correlations between audio features and a linear approximation of these using embeddings.

In Section~\ref{dim_reduc}, we present a technique of 2D visualization of these relationships.

\subsection{Style Classification score}
\label{score}

Obviously, embeddings computed from the style classification system were trained to have a high classification score. Here we investigate the suitability of the two other embedding types for a style classification task.  We measure their classification capacity in terms of %ANOVA F-value and
mutual information, as shown in Table~\ref{mi}.% and Table 3.
It measures the dependency between each of the 8 embedding dimensions and style categories.
Mutual information was computed with scikit-learn library~\cite{scikit-learn-11-pedregosa}.

%\textbf{Fengna: Obviously, embeddings computed from the style classifier provide the highest classification score, 94.38\% as stated in Section~\ref{style_classif}. However, we cannot feed the other two embeddings to the style classifier, instead, we measure their classification capacity in terms of %ANOVA F-value and
%mutual information, as shown in Table~\ref{mi}.% and Table 3.
%}

% \begin{table}[ht]
% \caption{ANOVA F-value between embedding dimensions and style}
% \label{anova}
% \begin{center}
% \begin{tabular}{|c|c|c|c|c|c|c|c|c|}
% \hline
% {} &  VAE-TTS &   Style &  Speaker \\
% \hline
% 0 &     2430 &   94692 &     1284 \\
% 1 &     6770 &   89641 &      746 \\
% 2 &    16015 &  208180 &     3139 \\
% 3 &     7037 &   75931 &     1533 \\
% 4 &    11456 &  107553 &      791 \\
% 5 &     9156 &   77273 &     1386 \\
% 6 &     9510 &  197631 &      828 \\
% 7 &    15468 &  160017 &      417 \\
% \hline
% \end{tabular}
% \end{center}
% \end{table}

%Je ne comprends pas les unités. Les f-sores sont souvent normés non?

\begin{table}[ht]
\caption{Mutual information (in bit) between embedding dimensions and style}
\label{mi}
\vspace{-5mm}
\begin{center}
\begin{tabular}{|c|c|c|c|c|c|c|c|c|}
\hline
{} &  VAE-TTS &  Style &  Speaker \\
\hline
0 &     0.44 &   1.55 &     0.41 \\
1 &     0.81 &   1.36 &     0.33 \\
2 &     1.08 &   1.49 &     0.50 \\
3 &     0.71 &   1.41 &     0.39 \\
4 &     0.97 &   1.47 &     0.33 \\
5 &     0.79 &   1.80 &     0.42 \\
6 &     0.97 &   1.74 &     0.25 \\
7 &     1.06 &   1.86 &     0.33 \\
\hline
\end{tabular}
\end{center}
\vspace{-5mm}
\end{table}

\subsection{Relationship between the Embedding Spaces and Audio Features}
\label{correlations}
In this analysis, we study the relationship between the embedding spaces and the eGeMAPS feature set~\cite{egemaps-16-eyben}. This feature set was designed based  on  their  potential  to  represent  affective  physiological  changes  in  speech.

This analysis allows to investigate what features describe best the remaining variability in the data.

The procedure is the following:
\begin{itemize}
    \item We approximate a linear function (i.e. hyperplane) between each latent space and the audio feature space with ordinary least squares linear regression. %We thus obtain a hyper-plan that approximate audio features from latent embeddings
    % je ne pige pas.
    \item We compute the linear function approximations of audio features based on latent embeddings
    \item The goal is then to evaluate how the approximations correlate with ground truth values. 
    \item To that aim, we compute the Absolute Pearson Correlation Coefficient (APCC) between predictions and ground truth
\end{itemize}

In other words, we compute the APCC between each audio feature and the best possible hyperplane, in terms of least squares, of each latent space.

To summarize these results, we present in Table~\ref{best_APCC} only features that have an $APCC>0.5$ in every latent space. 

\begin{table}[ht]
\caption{APCC values between the best possible hyper-plan of each latent space and audio features of the eGeMAPS feature set}
\label{best_APCC}
\vspace{-5mm}
\begin{center}
\begin{tabular}{|c|c|c|c|c|c|c|c|c|}
\hline
{APCC} &     VAE-TTS &     Style &     Speaker \\
\hline
F0 mean          &  0.76 &  0.82 &  0.63 \\
F0 percentile20.0 &  0.75 &  0.81 &  0.62 \\
F0 percentile50.0 &  0.79 &  0.86 &  0.67 \\
F0 percentile80.0 &  0.69 &  0.73 &  0.52 \\
mfcc2 mean                           &  0.73 &  0.77 &  0.65 \\
mfcc4 mean                           &  0.73 &  0.77 &  0.61 \\
F1 freq mean                   &  0.61 &  0.71 &  0.52 \\
F2 freq mean                   &  0.58 &  0.68 &  0.52 \\
F3 freq mean                   &  0.64 &  0.71 &  0.57 \\
Alpha Ratio V mean                   &  0.60 &  0.65 &  0.55 \\
Hammarberg Index V mean              &  0.58 &  0.63 &  0.52 \\
Slope V 0-500 mean                   &  0.89 &  0.91 &  0.72 \\
mfcc2 V mean                        &  0.78 &  0.82 &  0.68 \\
mfcc4 V mean                        &  0.77 &  0.80 &  0.63 \\
\hline
\end{tabular}
\end{center}
\end{table}

The feature set is based on Low-level descriptors (F0, formants, mfcc, etc.) to which are applied statistics for the utterance (mean, normalized standard deviation, percentiles).
All  functionals  are  applied  to  voiced  regions  only (non-zero F0). For MFCCs, there is also a version applied to all regions (voiced and unvoiced).

These features are defined in~\cite{egemaps-16-eyben} as follows:
\begin{itemize}
    \item F0: logarithmic F0 on  a  semitone  frequency scale, starting at 27.5 Hz (semitone 0)
    \item F1-3: Formants 1 to 3 centre frequencies
    \item Alpha  Ratio:  ratio  of  the  summed  energy  from 50-1000 Hz and 1-5 kHz
    \item Hammarberg Index: ratio of the strongest energy peak in the 0-2 kHz region to the strongest peak in the 2-5 kHz region.
    \item Spectral Slope 0-500 Hz and 500-1500 Hz: linear regression  slope  of  the  logarithmic  power  spectrum within the two given bands.
    \item mfcc1-4: Mel-Frequency  Cepstral  Coefficients 1 to 4
\end{itemize}

% \subsection{Selection of a feature subset}
% In the context of an interface for controllable speech synthesis, it would be useful to have a small number of features that gives a good overview. To extract a subset of the list, we investigate correlations between audio features themselves to exclude redundant data and select a subset. Figure~\ref{fig:corr_feat} shows the matrix of APCCs of audio features listed in Table~\ref{best_APCC}. Features with an APCCs $>=0.8$ are framed in red (except the diagonal which are 1 anyway).

% % show correlation matrix

% \begin{figure}[ht]
%   \centering
%   \includegraphics[width=\linewidth]{LaTeX/corr_feat_modif.png}
%   \caption{Correlations between audio features of Table~\ref{best_APCC}, the ones $>= 0.8$ are framed in red}
%   \label{fig:corr_feat}
% \end{figure}

% Based on this, we choose this subset of features:
% \begin{itemize}
%     \item F0 mean
%     \item F1 mean
%     \item mfcc 2, mfcc 4
%     \item alpha
%     \item slope
% \end{itemize}

\subsection{Dimensionality reduction of latent spaces}
\label{dim_reduc}

In this Section, we investigate the use of dimensionality reduction of latent spaces previously computed. Reducing the latent spaces to two dimensions will enable the possibility to design an interface that allows its visualization and its relationship with audio features.

To that aim, we use three different algorithms of dimensionality reduction: PCA, t-SNE and UMAP. We then perform the same procedure of regression as in Section~\ref{correlations} to obtain APCCs between each audio feature of Table~\ref{best_APCC} and the best possible hyper-plan, in terms of least squares, of each dimensionally reduced latent space.

\begin{figure}[ht]
%  \centering
  \includegraphics[width=1.15\linewidth]{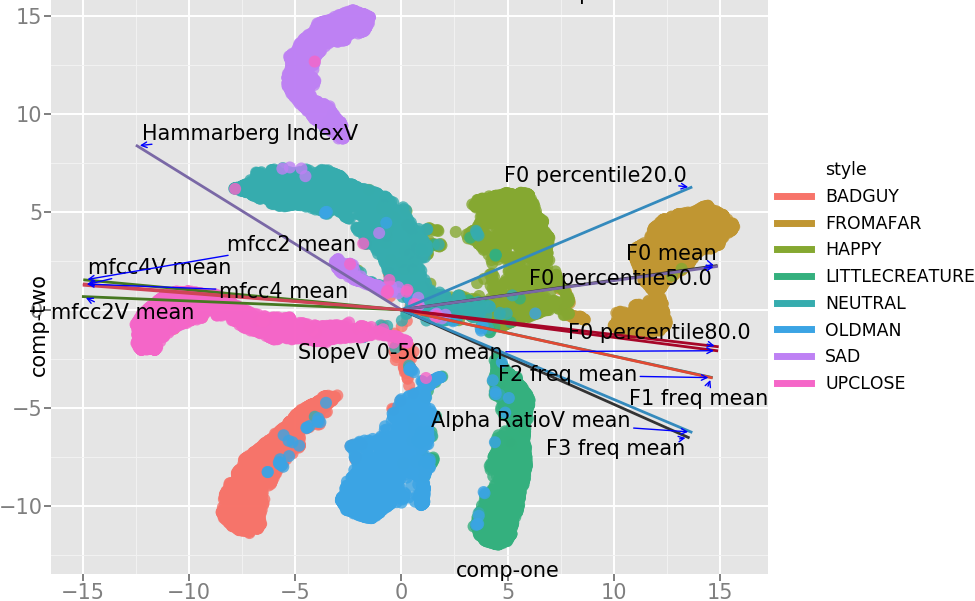}
  \caption{Latent space with directions of gradients of features listed in }
  \label{fig:gradients}
\end{figure}

Table~\ref{average_APCC_reduc} shows the average of APCCs for each pair (task, dimension reduction algorithm). For this pair, the gradients of hyper-plans approximating audio features were computed. The direction of these gradients correspond to the direction of the highest variation of a feature in the space. Figure~\ref{fig:gradients} shows the reduced embeddings of all utterances of the dataset and the directions of the gradients. 

This representation is useful for a perspective of interface for controllable speech synthesis system on which are represented the trends of audio features in the space.

\begin{table}[ht]
\caption{APCC average for each pair (task, dimensionality reduction algorithm)}
\label{average_APCC_reduc}
\vspace{-5mm}
\begin{center}
\begin{tabular}{|c|c|c|c|}
\hline
{APCC} &   PCA &  t-SNE &  UMAP \\
\hline
VAE-TTS &  0.564 &  0.422 &  \textbf{0.614} \\
Style   &  0.480 &  0.366 &  0.607 \\
Speaker &  0.512 &  0.480 &  0.549 \\
\hline
\end{tabular}
\end{center}
%\vspace{-10mm}
\end{table}

%\subsection{Discussion}

% Maybe explain that sooner (motivations ?)
%The reason we want to compute latent embedding spaces representing expressiveness instead of directly using audio features related to expressiveness is their usability in speech generation. We know that audio features are interdependent. Therefore, a control based on low-level features such as $F_0$ by analysis-synthesis will sound inconsistent. With the use of embeddings, we already know from the state of the art that we can generate consistent styled speech by varying their values. Thanks to the analysis of this work, we will also know how the embedding variables influence the audio features and be able to control it as we want.

% Discuss the link between these audio features and expressiveness: F0 and formants are obvious, and also the spectral shape corresponds to the vocal tension. Hammaberg and alpha are ratios of of energies high vs low, so probably also linked to vocal tension

%Embeddings based on style classification task have better scores as expected. 

%The advantage of the TTS unsupervised and the speaker classification techniques is that a modelling of style can be obtained without labels concerning the style. Compared to the speaker classification technique, the TTS unsupervised technique performs better in terms of correlation and mutual information with audio features and style classification.

\section{Conclusions}

This paper presents a methodology to build latent spaces related to style/emotion in speech and visualize it along with its relationships with important audio feature for a purpose of controllable speech synthesis.

To that aim, we compare three latent spaces computed by training deep learning-based systems on three different tasks. We then examined the potential of these latent spaces for style classification to confirm that they contain useful information for representing style.

We then studied the relationships between each latent space and audio features to obtain a sense of the impact of audio features on the expressed styles. This analysis consisted in an approximation of audio features from embeddings by linear regression. The accuracy of approximations was then evaluated in terms of correlations with ground truth. 

The gradient of these linear approximations were computed to extract the information of variations of audio features in speech. By visualizing these gradients along with the embeddings, we observe the trends of audio features in the latent space.

% mais tu ne dis pas trop quoi observer ni quelle conclusion tirer et d'après ton graphique, la plupart des données audio vont dans le meme sens (horizontal) dudes styles. Pas clair.

In the future, this representation will be used to control an expressive speech synthesis system.

%\section{Future Work}
%Build an interface for controlling the latent spaces  and use these embeddings to synthesize expressive speech.

% \subsubsection{Quantitative Results in Synthesis}
% The paper also proposes metrics for evaluating the prosody for the synthesis task:
% \begin{itemize}
%     \item Mel Cepstral Distorsion
%     \item Gross Pith Error
%     \item Voice Decision Error
%     \item F0 Frame Error
% \end{itemize}

\section{Acknowledgments}

%\thanks{No\'e Tits is funded through a PhD grant from the Fonds pour la Formation \`a la Recherche dans l'Industrie et l’Agriculture (FRIA), Belgium. }

No\'e Tits is funded through a PhD grant from the Fonds pour la Formation \`a la Recherche dans l'Industrie et l'Agriculture (FRIA), Belgium. 

Thanks to Acapela Group for providing the dataset, for the interesting insights and collaboration on embeddings computation.

% \subsection{Figures}

% All figures must be centered on the column (or page, if the figure spans both columns). Figure captions should follow each figure and have the format given in Figure~\ref{fig:speech_production}.

% \subsection{Tables}

% An example of a table is shown in Table~\ref{tab:example}. The caption text must be above the table.

% \begin{table}[th]
%   \caption{This is an example of a table}
%   \label{tab:example}
%   \centering
%   \begin{tabular}{ r@{}l  r }
%     \toprule
%     \multicolumn{2}{c}{\textbf{Ratio}} & 
%                                          \multicolumn{1}{c}{\textbf{Decibels}} \\
%     \midrule
%     $1$                       & $/10$ & $-20$~~~             \\
%     $1$                       & $/1$  & $0$~~~               \\
%     \bottomrule
%   \end{tabular}
  
% \end{table}

% \subsection{Equations}

% Equations should be placed on separate lines and numbered. Examples of equations are given below. Particularly,
% % 
% \begin{equation}
%   x(t) = s(f_\omega(t))
%   \label{eq1}
% \end{equation}
% % 
% where \(f_\omega(t)\) is a special warping function
% % 
% \begin{equation}
%   f_\omega(t) = \frac{1}{2 \pi j} \oint_C 
%   \frac{\nu^{-1k} \mathrm{d} \nu}
%   {(1-\beta\nu^{-1})(\nu^{-1}-\beta)}
%   \label{eq2}
% \end{equation}
% % 
% A residue theorem states that
% % 
% \begin{equation}
%   \oint_C F(z)\,\mathrm{d}z = 2 \pi j \sum_k \mathrm{Res}[F(z),p_k]
%   \label{eq3}
% \end{equation}
% % 

% \begin{figure}[t]
%   \centering
%   \includegraphics[width=\linewidth]{figure.pdf}
%   \caption{Schematic diagram of speech production.}
%   \label{fig:speech_production}
% \end{figure}

\bibliographystyle{IEEEtran}

\bibliography{refs}

% \begin{thebibliography}{9}
% \bibitem[1]{Davis80-COP}
%   S.\ B.\ Davis and P.\ Mermelstein,
%   ``Comparison of parametric representation for monosyllabic word recognition in continuously spoken sentences,''
%   \textit{IEEE Transactions on Acoustics, Speech and Signal Processing}, vol.~28, no.~4, pp.~357--366, 1980.
% \bibitem[2]{Rabiner89-ATO}
%   L.\ R.\ Rabiner,
%   ``A tutorial on hidden Markov models and selected applications in speech recognition,''
%   \textit{Proceedings of the IEEE}, vol.~77, no.~2, pp.~257-286, 1989.
% \bibitem[3]{Hastie09-TEO}
%   T.\ Hastie, R.\ Tibshirani, and J.\ Friedman,
%   \textit{The Elements of Statistical Learning -- Data Mining, Inference, and Prediction}.
%   New York: Springer, 2009.
% \bibitem[4]{YourName17-XXX}
%   F.\ Lastname1, F.\ Lastname2, and F.\ Lastname3,
%   ``Title of your INTERSPEECH 2019 publication,''
%   in \textit{Interspeech 2019 -- 20\textsuperscript{th} Annual Conference of the International Speech Communication Association, September 15-19, Graz, Austria, Proceedings, Proceedings}, 2019, pp.~100--104.
% \end{thebibliography}

\end{document}